\documentclass{article}
\usepackage[utf8]{inputenc}

\title{Mert Intern}
\author{ }

\usepackage{listings}
\usepackage{color}
\usepackage{booktabs}
\usepackage{multirow}
\usepackage{graphicx}

\definecolor{dkgreen}{rgb}{0,0.6,0}
\definecolor{gray}{rgb}{0.5,0.5,0.5}
\definecolor{mauve}{rgb}{0.58,0,0.82}

\begin{document}

\maketitle

\maketitle
\begin{abstract}
Task planning is an important component of traditional robotics systems enabling robots to compose fine grained skills to perform more complex tasks. Recent work building systems for translating natural language to executable actions for task completion in simulated embodied agents is focused on directly predicting low level action sequences that would be expected to be directly executable by a physical robot. In this work, we instead focus on predicting a higher level plan representation for one such embodied task completion dataset - TEACh, under the assumption that techniques for high-level plan prediction from natural language are expected to be more transferable to physical robot systems \sgella{minor: can we make this argument stronger?}. We demonstrate that better plans can be predicted using multimodal context, and that plan prediction and plan execution modules are likely dependent on each other and hence it may not be ideal to fully decouple them.  
Further, we benchmark execution of oracle plans to quantify the scope for improvement in plan prediction models.
\end{abstract}
\dilek{Do we talk about the plan annotations in this paper (i.e., how we got them and whether we will release them)?}
\aishwarya{There is a brief mention of it in the experiments section but there is nothing particularly special to release since all we do is filter the predicted action sequence to retain only object interaction actions.}
\sgella{adding a figure will make things clear on how we get GT plans and also in generall makes a lot of things clear in abstract/intro.}
\section{Introduction}

Task planning is an important component of traditional robotics systems enabling robots to compose fine grained skills to more complex tasks~\cite{kejia,oroplanning,openworldplanning}. Such robots typically have "skills" that abstract out a sequence of low-level motor control actions, for example, navigating from one point to another, picking up an object~\cite{bwibots}. Skills are typically parameterized, for example, a navigation skill is likely to have parameters representing the source and destination. A task planning module typically has some notion of the state of the environment in which the robot is operating in and notions of the expected state changes to that environment when a skill is performed. Then, given a goal whose completion requires the chaining of skills, the task planner can reason about a sequence of skills that if executed can complete the task~\cite{classicalplanning}. Typically when such a sequence is executed, there will be additional modules that detect whether the individual skills were executed successfully and re-plan if failures are detected~\cite{replanning}. We imagine our planning models presented in this paper as potential alternatives to classical planning in a future robotics system. In our case, we assume that a robot will have skills corresponding to various object interaction actions, which will additionally include navigating to the relevant object. While some of the skills used in this work, such as {\em slicing}, are not within the realm of current physical robots, we believe this is a useful level of abstraction at which a robot can learn to reason about task planning based on direct visual observations.   

There are limitations to exploring new modelling techniques in physical robots as physical robots tend to be slow and are more sensitive to damage caused by poor choices of actions~\cite{habitat}. As a result, simulated environments~\cite{habitat,ai2thor,Matterport3D} have become increasingly popular in recent years with a focus on training deep learning models from egocentric visual observations, and in some cases natural language instructions, to predict abstracted action sequences. 

\begin{figure*}[!h]
    \centering
    \includegraphics[width=\textwidth]{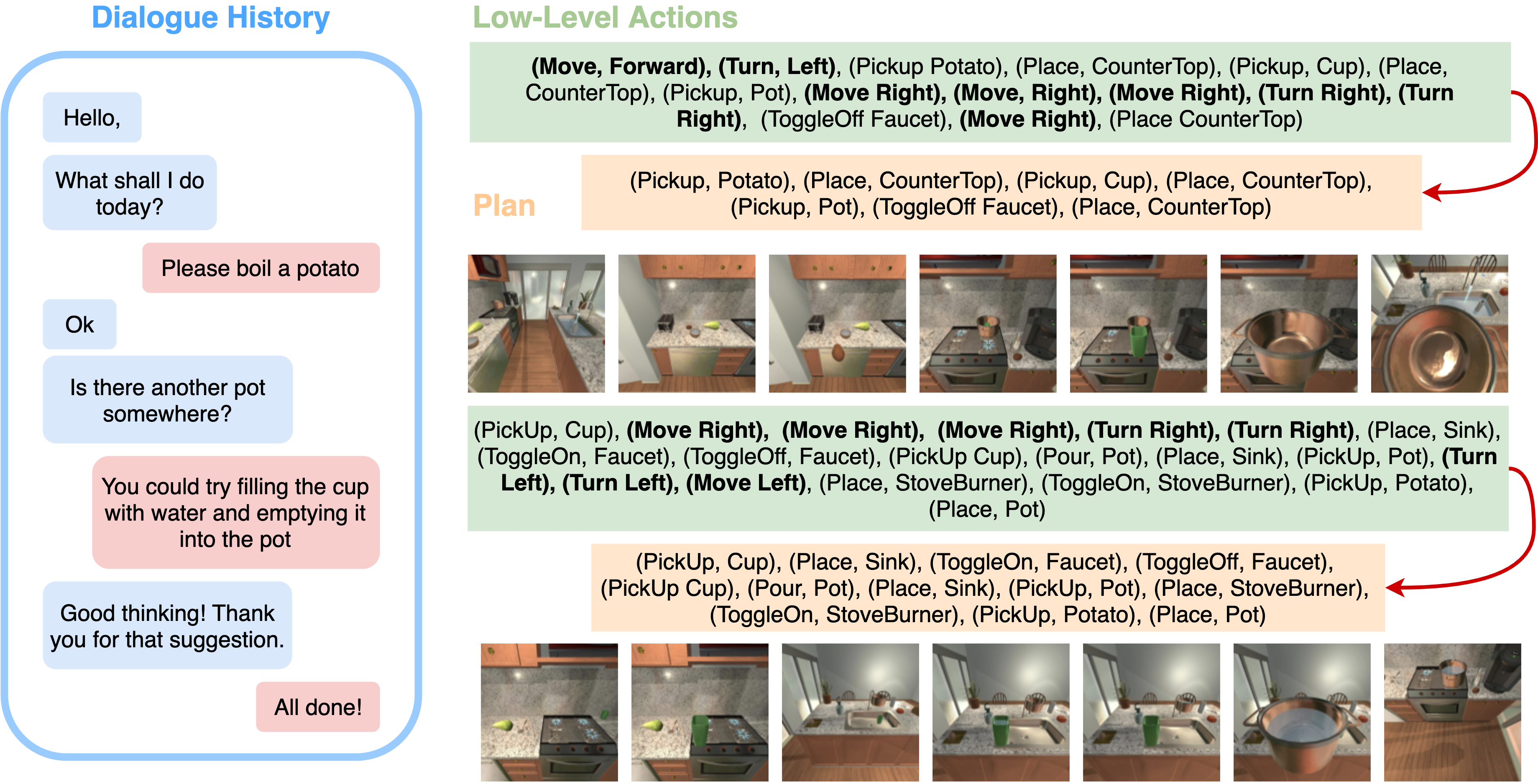}
    \caption{This figure depicts the main components of an EDH instance in the TEACh dataset. The dialogue history between the \textit{Commander} (red) and the \textit{Follower} (blue) happens concurrently with the low-level actions or the plan. The images are the history images of the driver robot in the simulator environment. For plan execution, all the low-level actions such as navigation are removed, resulting in a plan with high-level object interaction actions only.}
    \label{fig:edh_inst}
\end{figure*}
\sgella{decrease the size of the font in Figure 1? seems a bit big.}
\mert{made the font smaller.}

In this work, we focus on the \teach benchmark~\cite{teach}. This dataset consists of wizard-of-oz interactions between a user and an embodied agent collaborating via a natural language chat interface to complete household tasks. We focus on the Execution from Dialog History (\edh) task proposed for this dataset - given a segment of dialogue from the dataset as well as past actions and image observations, the model must predict subsequent actions in the environment to make progress on the relevant household task. The household tasks themselves are defined in terms of expected state changes in the environment and it is shown in the original \teach paper that it is difficult to develop a fully rule-based system to successfully execute the tasks~\cite{teach}. 

Prior work on other embodied task completion benchmarks have suggested that modular deep learning models with components for task planning, and semantic mapping and navigation, can outperform monolithic models that directly attempt to predict executable low-level actions based on language input and visual observations~\cite{film,amslam}. However, these results have been obtained on the \alfred benchmark~\cite{alfred} where task planning can be formulated as a simple classification problem. In contrast, the \teach\ dataset consists of more diverse tasks, including hierarchical and parameterized tasks and more diverse initial states, making planning non-trivial, as evidenced by the poor success rate of rule-based systems~\cite{teach}. 

In this work, we adapt the Episodic Transformer~\cite{et} (\et) model, originally proposed for predicting low level actions from multimodal observations for the ALFRED dataset to predict plans for the \teach \edh task. We show that in combination with a rule based plan execution module, this model produces plans that are significantly more successful than plans generated purely based on language information. We further explore modifications to the basic \et architecture that remedy problems observed with plans predicted by the \et model. We compare this performance to oracle plans obtained from human demonstrations in the original dataset, explore some limitations of our plan representation, and further analyze the execution of oracle plans to identify causes of failure in rule based execution of plans. We believe this analysis will provide insight that will better guide the development of benchmarks and models for embodied task completion.

\section{Related Work}

Task planning has long been a standard component of physical robot architectures~\cite{kejia}, particularly with general purpose service robots~\cite{bwibots,cobot}. Classical task planners include a symbolic representation of the state of the world, a goal and skills the robot is capable of executing, and are expected to find a sequence of skills that when executed will transform the world into the goal state, typically using heuristic search algorithms~\cite{classicalplanning}. Over the years, research in planning has improved the symbolic representations used in planners~\cite{strips,pddl,asp,10.5555/3241691.3241695,osti_10224709}, search algorithms~\cite{astar,causalgraphs,lamaplanner} and handling uncertainty via probabilistic methods~\cite{probplanning,planninguncertainty,ample}. More recent work has focused on expanding beyond fully defined world representations by expanding to use common sense~\cite{planningcommonsense} and open worlds~\cite{openworldplanning}.

A parallel track in recent years is work in simulated environments that focus on training deep learning models directly from egocentric visual observations instead of symbolic representations. Some simulators include Habitat~\cite{habitat, szot2021habitat}, AI2-THOR \cite{ai2thor}, Matterport-3D \cite{Matterport3D}, VirtualHome \cite{virtualhome,Liao_2019_CVPR,puig2020watchandhelp} and Chalet \cite{chalet}. Some tasks in such simulators are focused on learning sequences of discrete navigation actions including point goal navigation~\cite{pointgoalnav}, object goal navigation~\cite{objectnav}, or motor control for visual object manipulation~\cite{manipulathor}. Others involve both task planning and at least partial prediction of discrete navigation actions, for example transportation~\cite{threedworld} and rearrangement~\cite{housekeep}. More related to our work, vision and language navigation learning to convert natural language route instructions into sequences of discrete navigation actions~\cite{mattersim,touchdown,cvdn} and embodied task completion involves predicting a sequence of both discrete navigation and object manipulation actions based on natural language instructions~\cite{alfred,teach,cerealbar,arramon,mdc}. In this work, we adopt a modified version of the Execution from Dialog History (EDH) task defined in \cite{teach} and focus on predicting only the necessary object interaction actions based on language input and egocentric visual observations, ignoring navigation actions. 
\dilek{The related work can be organized more: for example, this second paragraph can move to the end or the beginning}
\aishwarya{I have sort of created the first paragraph as classical planning, the second as simulated environments and the third as planning in simulated environments. Not quite sure how to make it flow if I move this to either the beginning or end.}

There is some prior work focused on task planning in simulated environments for robots. \citet{alfredplanninggpt2} propose a model for task planning on the ALFRED dataset using GPT-2 which uses only the language input. Prior work has also explored task planning using only the language input in TEACh~\cite{teachda}. We use the BART model from this work as a baseline in our work. In contrast to \citet{teachda}, we explore the performance of multimodal planning models and evaluate based on success rate when combined with plan execution modules instead of relying on surface level metrics comparing ground truth and predicted plans. Some end-to-end models for the ALFRED dataset also have task planning modules as a component of their architecture~\cite{film,amslam}. However, due to the way the ALFRED dataset was created, task planning in ALFRED can be cast as a simple 7-way classification problem, whereas task plans for TEACh are more complex and often are very dependent on the initial state of the environment.  

\section{Task Setup}

The TEACh dataset \cite{teach} is a situated dialogue corpus, where two interlocutors communicate over a chatting interface simulating the interaction between a user (\commander) and embodied agent (\follower) to complete household tasks. The \commander\ has access to the task steps and the location information of objects, while only the \follower\ can interact with objects in the environment. The \follower\ has an egocentric view of the environment and can pose questions to the \commander\ about the location and task information. The \follower\ and must use this rich lexical and visual context to complete household tasks that require chaining long sequences of actions and reasoning about physical state changes.

In this paper, we focus on the Execution from Dialog History (EDH) benchmark on the TEACh dataset. Given past dialogue and actions from a human-human gameplay session, a model must predict subsequent actions the \follower\ would take in the environment until the next dialogue action. The model is evaluated by comparing the object state changes caused by the predicted action sequence with state changes caused by the ground truth action sequence. The action sequence predicted by the model is expected to be directly executable in the \teach simulator. This action space includes discrete actions for navigation - \texttt{Move Forward}, \texttt{Turn Right}, \texttt{Turn Left}, \texttt{Move Backward}, \texttt{Strafe Left}, \texttt{Strafe Right}, \texttt{Look Up} and \texttt{Look Down} - as well as object interaction actions - \texttt{Pickup}, \texttt{Place}, \texttt{Open}, \texttt{Close}, \texttt{Toggle On}, \texttt{Toggle Off}, \texttt{Slice}, \texttt{Pour}. For object interaction actions, models must predict a relative $(x, y)$ coordinate on the last egocentric image observation that are resolved by the underlying simulator into the object the action is to be executed upon. Using this, current neural models have at most a 10\% success rate on the \teach EDH task~\cite{teach} \sgella{I thought the current state-of-the-art is more than 10\%?}.
\mert{In the TEACh paper it shows aroun 10\%}

In contrast, in this paper, we modify the expected prediction from a model to be a "plan", which we define as being a sequence of object interaction actions paired with the object category of the object they are to be executed upon. 
An example for the task of boiling a potato is included in Figure \ref{fig:edh_inst}. 
We can see that the original low level action sequence includes navigation actions for completing many object interaction actions. In order to pick up a potato, the agent must navigate near it and potentially alter where it is looking so that the potato is in view. 
However, when predicting a plan, a model can simply predict that the agent needs to pick up a potato and then a separate plan execution module (heuristic in our case but could be learned) positions the agents in the correct location to perform the action.  
While it is possible to define plans at an even higher level, say having subgoals such as filling a cup with water which then need to be broken down, we choose the level of object interaction actions as our chosen level of abstraction as plans of this level can be automatically created from ground truth actions sequences eliminating the need for annotation of plans. 

Since the EDH task involves continuing a partially completed session, a particular EDH instance may show from the dialog that for example the cup is already filled with water. In this case, the plan the model would have to predict for this instance may only involve the last steps of pouring the water into the pot, setting it on the stove and adding the potato. 

With many \teach tasks, the task may also be parameterized, in which case plan prediction additionally involves identifying task parameters based on the dialog. For example, in the task of making breakfast, parameters are used to determine which dishes are to be prepared and how many of each are required. Changes in parameters change the plan the model is expected to predict, making the plan prediction problem less trivial than in earlier embodied task completion benchmarks. Further, the \teach dataset includes a more diverse and difficult set of initial states, requiring reasoning to produce object interaction sequences to say open a cabinet to take out a knife from inside it or clearing out the sink to place a large plate.     
\dilek{It would be useful to include more details in the appendix, such as full action sets both for low level actions and plans, an example with annotations of two types, etc. I think an overview figure would also be useful here (either an example session or model figure or both)} 
\aishwarya{The action space is kind of mentioned above but I'll see how to make it clearer. Mert is working on a figure which might help.}
\sgella{try to refer to the figure added}
\mert{added reference in the beginning.}

During inference, at each time step, our plan prediction model is expected to predict one object interaction action and the corresponding object category for the object on which it is to be executed. This plan step is then executed by one of two possible plan execution modules described in section~\ref{sec:exec}. Execution terminates either when a model predicts a special \texttt{Stop} action or if the model predicts 30 plan steps that result in an execution failure from the simulator, or reaches a limit of 100 plan steps. A plan step may fail execution for a variety of reasons. It may be infeasible, for example, trying to pick up a cabinet, a prerequisite step may not be completed (for example the \texttt{Slice} action is only feasible if the agent is holding a knife, or the execution module may not be able to find a feasible position for the agent from which to execute the action, for example the agent may not be positioned correctly in order to see and hence act on an object inside a drawer.  

\begin{figure}[!ht]
    \centering
    \includegraphics[width=8cm,trim={0 0 0.05cm 0},clip]{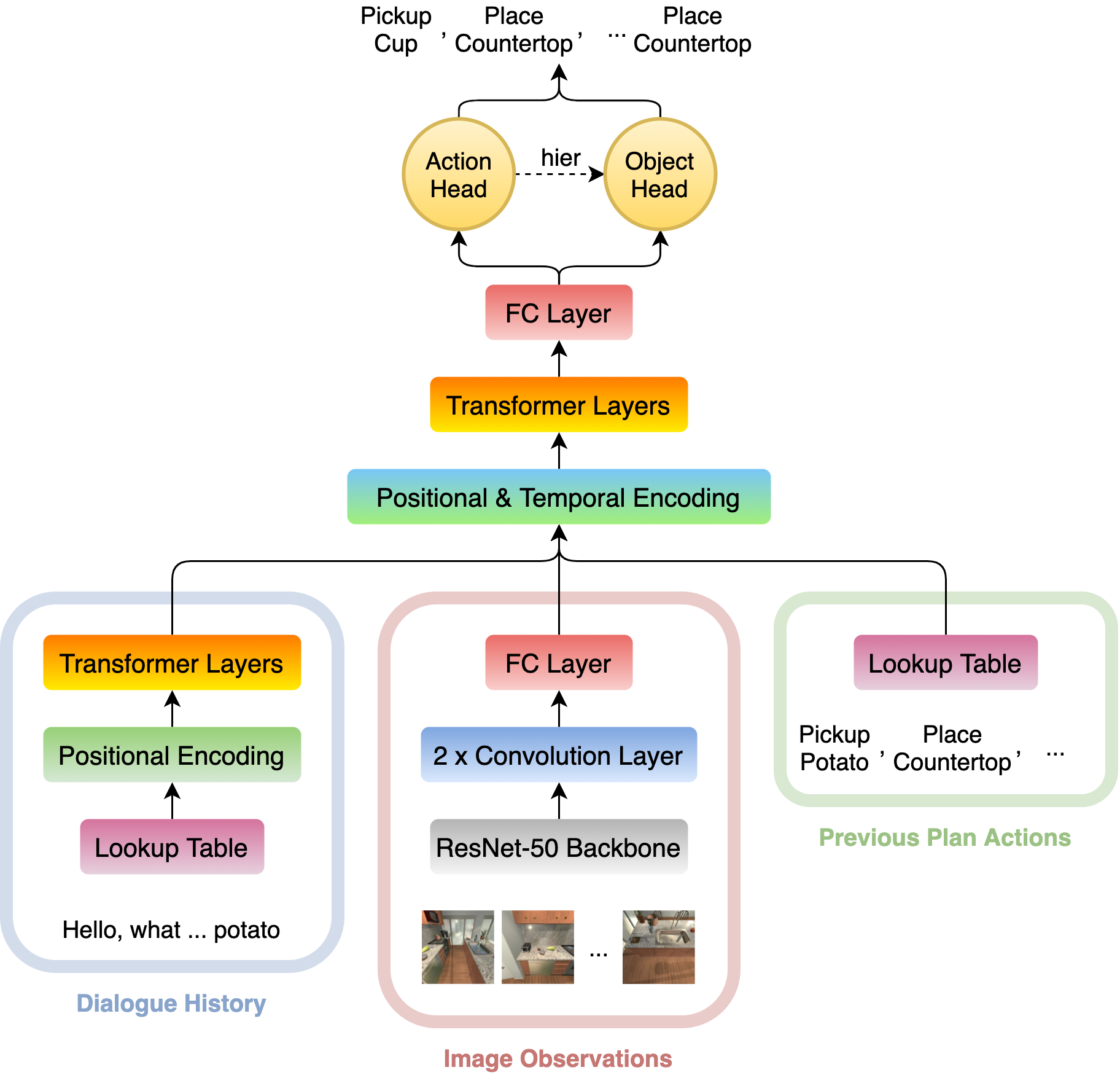}
    
    \caption{This is a depiction of the architecture for the \et-based models. The basic E.T. model does not have a connection between the action and object heads, but \hier ~does. There are three main input processing components (dialogue history, image observation and previous plan actions). These are then input into the positional and temporal encodings and transformer layers to come up with the next feasible set of high level plan actions.}
    \label{fig:model}
\end{figure}
\sgella{figure is taking up too much space here. I made it single column. Feel free to revert it back if you are not happy}
\mert{Thank you! I was thinking about making it smaller as well but couldn't decide. I think it may be better to make it slightly bigger as the font is not very readable right now and it is smaller than the smallest script size available i think which is not allowed on the guidelines.}
\mert{I made it ever so slightly bigger.}

\section{Plan Prediction Models}
\label{sec:models}
We modify the Episodic Transformer (\et) model~\cite{et}, which has previously been used for action prediction in ALFRED as well as \teach~\cite{teach}, for plan prediction. The E.T. model consists of a transformer encoder for language input (which in our case is the EDH dialog history) and image observations encoded using a ResNet-50 backbone which are concatenated and passed through multimodal transformer layers and then converted via linear layers into action and object category predictions. The model is designed to predict as many actions and objects as there are image observations. At train time, the model receives image observations for the entire trajectory and is expected to produce actions for the entire trajectory. At inference time, the model receives image observations for the actions completed so far in the trajectory, as well as past actions as input, and the last action in its predicted sequence is used as the predicted action for the next time step. To modify the model for plan prediction instead of low level action prediction, we modify the training data retaining only the object interaction objects and image observations from the corresponding time steps. During inference, after each plan step is executed, the last image observation from this execution is appended to the visual input to obtain the next plan step.

We explore the following variants of the \et model:
\begin{itemize}
    \item \textbf{\et}: Basic \et model described above.
    \item \textbf{\hier}: The \et model has independent classification heads predicting the action and object for each time step - occasionally resulting in infeasible actions such as picking up a cabinet. Instead of having a local classifier per level, we explore sharing information between the two heads by concatenating the output of the action classifier head with the input of the object classifier head. This is hypothesized to be giving more information to the higher level object classifier to select the action based on action-object pair validity. The model is hypothesized to learn valid object-action pair patterns from valid samples, yet, this is not fully enforced wit a hard constraint on the classifier. This may still lead to invalid object-action pairings.
    \item \textbf{\mask}: As an alternative to hierarchical learning, during inference, we explore whether the predicted action and object form an executable plan step. If not, we replace the action with the next most probable action that can be executed on the predicted object~\footnote{We modify the action rather than the object as we expect object prediction to be easier as objects are visible in the image observations.}. \sgella{do we have evidence on this that object prediction is easier than actions? may be from baseline model?} \aishwarya{Not really. It is kind of hard to figure out how to evaluate something like that since once the predicted plan diverges from the ground truth plan, it's hard to say whether the object or action is correct.}
\end{itemize}

\section{Plan Execution}
\label{sec:exec}

In this work we focus on modeling plan prediction. While predicted plans can be compared to ground truth plans, the best measure of whether a plan is correct is whether executing it completes the task as expected. To evaluate our predicted plans in this manner, we pair them with two possible rule based plan execution modules. We plan to explore the task of developing machine learning models for plan execution in future work.

\subsection{Direct Plan Execution}

Given a plan step consisting of an action and associated object type, we need to identify the particular object the agent must manipulate. To do this, we use the heuristic of selecting the object of the desired type closest to the agent. We then use the navigation graph from the simulator to compute the shortest path to the object, navigate to it and attempt the predicted action. 

\subsection{Assisted Execution}

Direct plan execution can fail for a variety of reasons. For instance, if the sink is full, and the predicted plan step requires placing a new item there, we need to empty it. 
While with our proposed level of abstraction we expect the plan prediction model to predict the necessary object interaction actions for this, we are also interested in seeing whether a model can be more successful if some of these details are abstracted out.
During assisted execution, we identify common such situations that cause failures and execute additional actions to address the situation and increase the likelihood that the plan step can be successfully executed. 


Specifically, we add the following assistive steps:
\begin{itemize}
    \item For all actions, if the target object property change is already complete, do nothing to avoid an execution failure. 
    \item \textit{Pickup:} If the object is inside a receptacle (container), open the receptacle. After pickup, if a receptacle was opened, close it.
    \item \textit{Place:} If the target receptacle is in a receptacle, take it out and place it on the counter first (for example, if we need to place something on a plate which is inside a drawer). If the target receptacle needs to be opened, open it and close after placement (for example, a drawer or microwave needs to be opened to place something inside). If a placement attempt fails, try removing existing contents of the receptacle one by one to make more space.
    \item \textit{Open, Close:} Toggle off target object if relevant (for example, microwaves need to be turned off to open them). 
    \item \textit{ToggleOn, ToggleOff:} If the target is open, close it first (for example, microwaves need to be closed to turn them on).
    \item \textit{Slice:} If the target is in a receptacle, first move it to the counter.
\end{itemize}
Additionally, we also attempt position adjustments to increase the chance of success.  \sgella{so do you consider this or not in models presented here? - asking since in results error analysis you mention about a lot of placement errors. So it would be helpful to elaborate on it or at least add a footnote.} \aishwarya{Bit confused. I thought the results section says that position errors are rare?}


\begin{table*}[t]
    \centering
    \begin{tabular}{ll|rrrr|rrrr}
         \toprule
         && \multicolumn{4}{c}{\textbf{EDH Divided Val Split}} & \multicolumn{4}{c}{\textbf{EDH Divided Test Split}} \\
         && \multicolumn{2}{c}{\textit{Seen}} & \multicolumn{2}{c}{\textit{Unseen}} & \multicolumn{2}{c}{\textit{Seen}} & \multicolumn{2}{c}{\textit{Unseen}} \\
         Model & Execution & \multicolumn{1}{c}{SR} & \multicolumn{1}{c}{GC} & \multicolumn{1}{c}{SR} & \multicolumn{1}{c}{GC} & \multicolumn{1}{c}{SR} & \multicolumn{1}{c}{GC} & \multicolumn{1}{c}{SR} & \multicolumn{1}{c}{GC} \\
         \midrule
\multirow{2}{*}{Baseline} & Direct & 11.26 & 13.67 & 7.51 & 11.03 & 7.19 & 9.62 & 8.87 & 9.54\\
 & Assisted & 11.92 & 17.27 & 8.91 & 12.19 & 9.80 & 12.30 & 10.27 & 12.31\\
\multirow{2}{*}{E.T.} & Direct & 12.91 & 16.32 & 15.58 & 16.20 & 15.03 & 19.52 & 16.62 & 15.61\\
 & Assisted & 15.89 & 20.57 & 18.74 & 22.36 & 16.67 & 19.96 & 19.98 & 27.13\\
\multirow{2}{*}{E.T. Hierarchical} & Direct & 14.24 & 15.67 & 16.23 & 17.27 & 14.71 & 17.97 & 17.27 & 20.30\\
 & Assisted & 18.21 & 20.45 & 18.09 & 24.53 & 17.97 & 23.67 & 19.70 & 25.82\\
\multirow{2}{*}{E.T. + Mask} & Direct & 15.23 & 22.51 & 17.81 & 18.29 & 16.34 & 23.84 & 17.46 & 18.96\\
 & Assisted & 18.87 & 28.99 & 19.57 & 27.64 & 18.95 & 26.35 & 20.07 & 28.33\\
\multirow{2}{*}{Oracle} & Direct & 61.92 & 63.64 & 55.57 & 58.48 & 54.58 & 53.58 & 56.77 & 58.01\\
 & Assisted & 68.87 & 72.13 & 61.97 & 63.07 & 61.44 & 62.49 & 63.21 & 64.87\\
\multirow{2}{*}{CorefOracle} & Direct & 77.81 & 83.03 & 70.87 & 71.87 & 75.82 & 79.50 & 71.90 & 74.34\\
 & Assisted & 80.13 & 84.50 & 74.58 & 77.31 & 78.43 & 80.92 & 76.94 & 78.30\\
    \bottomrule
    \end{tabular}
    \caption{Success rate (SR) and Goal Condition Success Rate (GC) of different models combined with different execution methods on the \teach\ EDH task. Oracle performances are separated as upper bounds on the task. Best performance results are bolded for each metric and split in the specific execution method.}
    \label{tab:success}
\end{table*}

\begin{table*}[!ht]
    \centering
    \tabcolsep 2pt
    \begin{tabular}{L{2cm}L{2cm}|C{1.5cm}C{2cm}C{2cm}C{2cm}}
        \toprule
        Model & Execution & Edit Distance & Frac Valid Plan Steps & GT-Norm ED & Pred-Norm ED \\
        \midrule
\multirow{2}{2cm}{Baseline} & Direct & 4.87 & 94.61 & 1.39 & 1.22\\
 & Assisted & 4.87 & 94.61 & 1.39 & 1.22\\
\multirow{2}{2cm}{E.T.} & Direct & 45.78 & 90.33 & 21.72 & 0.96\\
 & Assisted & 54.80 & 89.08 & 25.48 & 0.96\\
\multirow{2}{2cm}{E.T. Hierarchical} & Direct & 45.08 & 88.25 & 21.49 & 0.96\\
 & Assisted & 51.52 & 86.84 & 23.75 & 0.95\\
\multirow{2}{2cm}{E.T. + Mask} & Direct & 49.43 & 98.51 & 22.82 & 0.96\\
 & Assisted & 60.00 & 98.69 & 27.51 & 0.96\\
        \bottomrule
    \end{tabular}
    \caption{Edit distance, fraction of valid plan steps, and Ground Truth (GT) and prediction (pred) length normalized edit distance (ED) of different models combined with different execution methods on the \teach\ EDH \texttt{divided\_valid\_unseen} split. Oracles are not included here as by definition their plans are always valid and have edit distance 0.}
    \label{tab:all_metrics_div_val_unseen}
\end{table*}

\section{Experiments}
We evaluate our proposed plan prediction models on the EDH task of the TEACh dataset~\cite{teach}. We experiment with each of the models in section \ref{sec:models} with each execution method in section \ref{sec:exec}. Additionally, we evaluate the following baseline and oracle conditions:
    \paragraph{\texttt{Baseline}:} Our baseline is a BART model, original proposed in \cite{teachda}, that uses only language input--the EDH dialogue history--and predicts the entire plan as a sequence. As a result, it neither accounts for specific aspects of the environment that are not explicitly discussed, nor can it adapt to account for plan steps that failed to execute. 
    \paragraph{\gt:} As an upper bound to the success rate obtainable with each of our plan execution methods, we obtain oracle plans using the ground truth actions present in the EDH instance. We filter these actions sequences retaining only object interaction steps and converting object IDs to object types to match the plan representation used by our models.
    \paragraph{Oracle with Object IDs (\gtcoref):} To further test the extent to which our plan representation combined with our heuristic of selecting the closest object of a particular type limits performance, we also evaluate another oracle that we call \gtcoref. This produces plans containing object IDs instead of object types so that during plan execution, the execution module knows which object of a particular type must be manipulated. 
We do not compare to the TEACh EDH baselines in this paper as our execution methods access information that the TEACh baseline models are not allowed to access. However, all our models except the baseline have higher success rates than the TEACh baselines.

We use the following metrics for evaluation:
    \paragraph{Edit distance:} between ground truth and predicted plans, considering plan steps to have an edit distance of 1 if they match both in terms of the action and object and 0 otherwise~\footnote{See appendix for an example.}.
    \paragraph{Normalized Edit distance:} Additionally, by definition, edit distance depends on the original length of the ground truth or predicted plans. As the length of the plan increases, the possibility of having a larger edit distance increases. This may be misleading in certain cases when the distribution of the lengths of the plans is skewed~\footnote{see appendix for detailed analysis of the skewed distribution.}. Hence, we explore two length-normalized edit distances to give more insight into the tuple differences. 
\begin{itemize}
    \item Ground-Truth-Length Normalized Edit Distance = $\frac{Edit Distance}{Length of the Ground Truth}$
    \item Prediction-Length Normalized Edit Distance = $\frac{Edit Distance}{Length of the Predicted Plan}$
\end{itemize}.
    \paragraph{Fraction of valid plan steps:} We consider a plan step to be valid if it is possible to execute the predicted action on an object of the predicted object type. For example \texttt{(Pickup, Potato)} would be valid while \texttt{(Pickup, Sink)} would not.
    \paragraph{\teach\ execution metrics:}
    \sgella{you can skip elaborating on \teach metrics here and refer to \teach paper. Saves some space.}
    \aishwarya{True but it seems over-ambitious to cut down to 4 pages and we are still below 8 now.}
    \begin{itemize}
        \item Success Rate (SR): Fraction of successful EDH instances, that is, EDH instances for which all desired object state changes occurred on executing the predicted plan.
        \item Goal Condition Success Rate (GC): Fraction of desired object state changes across EDH instances that were completed through execution of predicted plans.
    \end{itemize}

Since the \teach\ test set is not public, we follow the standard protocol proposed in the \teach\ codebase~\footnote{\url{https://github.com/alexa/teach}} of using a standardized division of the original validation sets into validation and test sets called the divided validation and divided test sets. Additionally, both for validation and testing, there are EDH instances situated in the same floorplans as training instances (seen), and those situated in completely new floorplans (unseen). The additional challenge imposed by unseen environments are visual differences with respect to the training data. 

\section{Results}
\sgella{add some task level analysis or insights to understand which tasks (probably the ones that has on average less object manipulations or issues with positions is doing better? And does the trend look similar to \teach sota edh metrics (if not why?). what does mean? navigation is a issue vs. object manipulation etc. Some discussion like that?}
\aishwarya{Thanks! This now exists in appendix. Working out best way to add some of the insights here. We did not directly compare to the TEACh baseline because they are not working in an identical setting (stopping conditions are different)}
We present \teach\ execution metrics obtained from different models paired with different plan execution conditions in table \ref{tab:success}. This additionally includes results from the two oracle models. For a subset of these conditions we train and perform inference with 3 random seeds and perform 2 sided Welch t-tests. Allowing for Bonferroni corrections over 4 tests, we find that \mask ~is trending to be significantly better than the baseline with $p = 0.0381$ on the \texttt{divided\_val\_seen} split and $p = 0.0164$ on the \texttt{divided\_test\_seen} split. We did not find any statistically significant different between the \hier ~and \mask ~models~\footnote{We did not perform statistical comparisons across all pairs of conditions as it is expensive and time consuming to run inference  with enough random seeds to allow for Bonferroni corrections as the number of tests grows.}.  

We observe that even with oracle plans combined with assisted plan execution, the success rate is not 100\%. 
The highest oracle success rate is achieved using object IDs (\gtcoref) in assisted plan execution with 80.13\% for validation and 78.43\% for test splits. Our regular \gt\ model sharing the same plan representation as our proposed models has a success rate of 68.87\% for validation and 61.44\% for test sets. 
This gap suggests that learning to disambiguate which particular instance of an object needs to be manipulated in each plan step plays a substantial role in eventual success rate. Despite using heuristic plan execution, for both oracle conditions we observe an unexpected drop in scores from seen to unseen splits. Further, even with oracle plans, assisted execution plays an important role in increasing success suggesting that low level position and placement adjustments play an important role in overall task success. 
Additionally, for all the cases in oracle performance, goal condition success rate is slightly higher than the regular success rate suggesting that even when plan execution fails for some EDH instances, enough progress is made to complete at least some of the desired state changes.

When examining models, the baseline starts with a success rate of around 11\%. There is a drop in success rate from seen to unseen environments and assisted execution is primarily beneficial in unseen environments. 
The results for the \et, \hier\ and \mask\ improve over the baseline but have a long way to go to meet even \gt\ which uses the same plan representation. 
We see that the best performing model is \mask\ in most cases. We conjecture that this is due to the fact that we enforce a hard constraint on the invalid action-object tuples so that the model never is able to predict invalid tuples such as \texttt{(Pickup, Sink)}. This gives a performance improvement compared to the baseline. 

Further, while assisted execution improves success rate in most cases, the improvements for models are much smaller than the \gt\ condition. We find that in some cases the trends between the various \et\ models in direct execution do not match those in assisted execution. For example, in the \texttt{divided$\_$test$\_$seen} split, \et\ outperforms \hier\ on direct execution but the reverse trend is observed with assistance. This suggests that the plan execution mechanism may affect the quality of plan prediction, at least in models like ours where the observations from executing one plan step are used in model input when obtaining the next plan step.

In the appendix, we also examine a breakdown of the success rate by task. We find that the oracle conditions have a lower success rate in tasks that either involve more task steps or more complex placement actions, although there is still a considerable gap between model and oracle performance. 
For the \et\ models, we see consistent increases in task level success rate for most tasks from \et\ to \hier\ and \et\ to \mask\ but the trend between \hier\ and \mask\ is less consistent. The baseline outperforms vanilla \et\ on some tasks but almost never outperforms \hier\ or \mask. 
From qualitatively comparing the predictions from the \bart and the various \et models, we can see that the multimodal input is primarily beneficial in understanding how much of the task specified in the dialog history has already been completed in the action history. Plans generated by the \bart\ model sometimes simply repeat a few plan steps from the history and fail to proceed further. Additionally, the multimodal input is able to help when the dialog history does not lay out detailed instructions on how a task is to be accomplished. For example, in the \texttt{Water Plant} task, the agent usually needs to fill some smaller container like a cup with water from the sink and pour this into the part. The \bart model generates correct plans if the dialog history explicitly describes this process but not if it just involves the \commander\ asking the \follower\ to water the plant. However, the \et\ model appears to get prompted by the objects in the scene and is able to solve the task. Such visual prompting can also have the opposite effect though where in some cases the \et\ model ignores the language input and performs unrelated manipulations on easily visible objects, or ignores small objects in favour of larger, easier to see objects. 

We additionally compare the performance of models using edit distance and fraction of valid plan steps on the \texttt{divided$\_$val$\_$unseen} split in Table~\ref{tab:all_metrics_div_val_unseen}. In contrast to success rates we observe that the baseline has a much lower edit distance of around 5 compared to the \et\ models at around 45 for direct execution. On further inspection, we find that this is because the \et\ models do not learn to predict when to stop. As such the plan ends when they hit the limit of plan step execution failures, which typically extends much longer than the ground truth plans. Their edit distance goes up with assisted execution further contrary to success rate as better execution means that more plan steps can be executed before reaching the limit of plan execution failures. We additionally find that the \et\ model starts out predicting more invalid plan steps than the baseline but this is fully corrected in \mask\ by design. One counterintuitive observation is that the \hier\ model, which was also designed to reduce invalid plan steps, is found to have the opposite effect.

Finally we examine the causes of plan execution failures of oracle plans, particularly the \gtcoref. We find that navigation failures are rare, accounting for less than 1\% of the failure cases. We find that placement is particularly challenging, particularly when an object needs to be placed on another object that likely already contains other objects. We find that placing objects on a stove always fails, and placing items on dressers and coffee tables fails more than 50\% of the time they are attempted. Opening and closing of cabinets and microwaves are also found to be challenging with failures primarily occurring because of fine adjustments to the agent's position that are difficult to calibrate heuristically.   
\subsection{Conclusion}

We develop a model for multi-modal plan prediction for the TEACh dataset using the Episodic Transformer architecture and evaluate end to end performance on the TEACh EDH task in conjunction with heuristic plan execution modules.
Our \et\ plan prediction models outperform a language only baseline but are quite far from oracle performance. The main cause of failure of our models is failing to accurately predict when to stop.
We find that programmatic assistance in plan execution improves EDH success rates by a large margin on oracle plans but only to a limited extent on model generated plans.. 
We also find that plan model improvements do not maintain the same trend with different plan execution modules suggesting a dynamic interaction between plan prediction and plan execution improvements.
Our work suggests opportunities for future work in plan representation, prediction and execution - as we show that our heuristic of navigating to the closest object substantially reduces the success rate of executing oracle plans, there is a significant gap between model and human generated plans, and our heuristic plan execution module is still only successful at executing human plans about 80\% of the time.

\bibliography{anthology,custom}

\appendix

\section{Training Hyperparameters}

For training the \et, \mask\ and \hier\ methods, we retained hyperparameters from the original TEACh paper~\cite{teach}, except the batch size, without further hyperparameter tuning, and used the largest batch size that could fit in a single GPU of a p3.8xlarge AWS EC2 instance.
We used the AdamW optimizer with 0.33 weight decay with a learning rate of 1e-4 for the first 10 epochs and 1e-5 for the last 10 epochs. We trained all models for 20 epochs with a batch size of 3, and report results from the final epoch. 
We used two transformer layers for the language encoder with 12 attention heads and an embedding size of 768, and 2 multimodal transformer layers with 12 attention heads. 
We replace sampling with rotation permutations of our training dataset per epoch, ensuring that every train example is seen exactly once in our dataset. For language decoder in the transformer we use a drop-out of 0.1, and for the encoder we use a dropout of 0.1.
The different \et\ models required 4 hours for preprocessing (extracting image features using the ResNet-50 backbone), about 2 hours per model for training using 4 GPUs of a p3.8xlarge AWS EC2 instance. At inference time we could use a maximum of 3 GPUs for inference as one GPU was required by the simulator. When using 3 GPUs of a p3.8xlarge AWS EC2 instance, E.T. models took about 11 hours to complete inference jointly on the \texttt{divided\_val\_seen} and \texttt{divided\_test\_seen} splits and about 35 hours to complete inference jointly on the \texttt{divided\_val\_unseen} and \texttt{divided\_test\_unseen} splits. The time difference is due to the size of the various splits. 

For the baseline BART model, we retain hyperparameters from the model presented in \cite{teachda}. We take the pretrained BART-base model from the Huggingface library~\footnote{https://huggingface.co/} and finetune for 20 epochs using a batch size of 2 per GPU. Training was done using gradient accumulation across 4 GPUs of an p3.8xlarge AWS EC2 instance. We use the AdamW optimizer with $\beta_1 = 0.9$, $\beta_2 = 0.99$, $\epsilon = 1e-08$ and weight
decay of 0.01. We use a learning rate of 5e-05 with a linear warmup over 500 steps. 
The BART model can be finetuned in under an hour using all 4 GPUs of a p3.8xlarge AWS EC2 instance. We first performed inference on the BART model and saved the predicted plans to file before separately executing them in the AI2-THOR simulator. This process can also be completed in under an hour. 
Executing stored plans either in the case of the BART model or the oracle conditions took about 2.5 hours using 3 GPUs of a p3.8xlarge AWS EC2 instance for the combined \texttt{divided\_val\_seen} and \texttt{divided\_test\_seen} splits and about 8 hours for the combined \texttt{divided\_val\_unseen} and \texttt{divided\_test\_unseen} splits.

\section{Edit Distance}

\begin{figure*}[!h]
    \centering
    \begin{tabular}{cc}
        \includegraphics[scale=0.5]{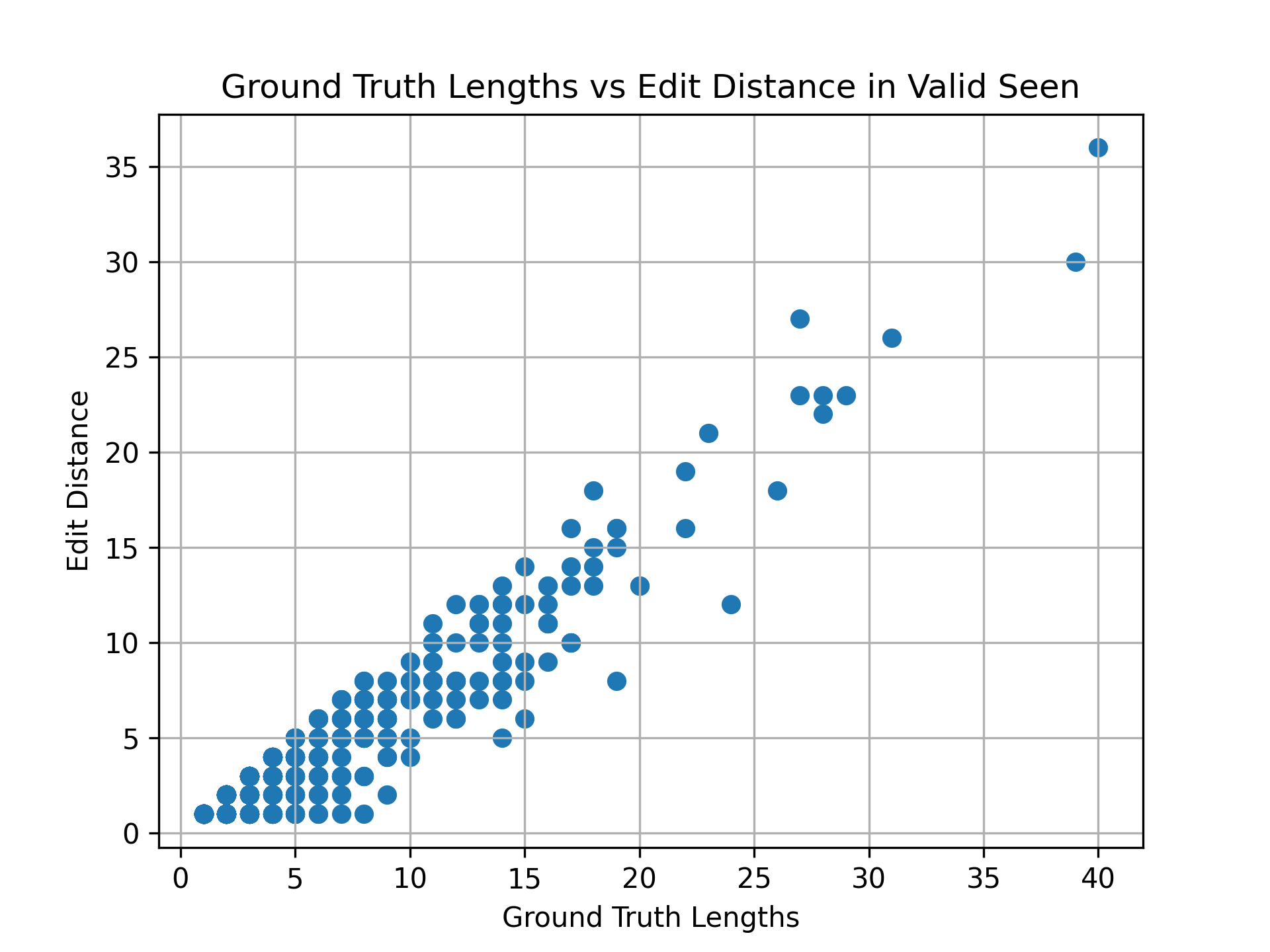} & \includegraphics[scale=0.5]{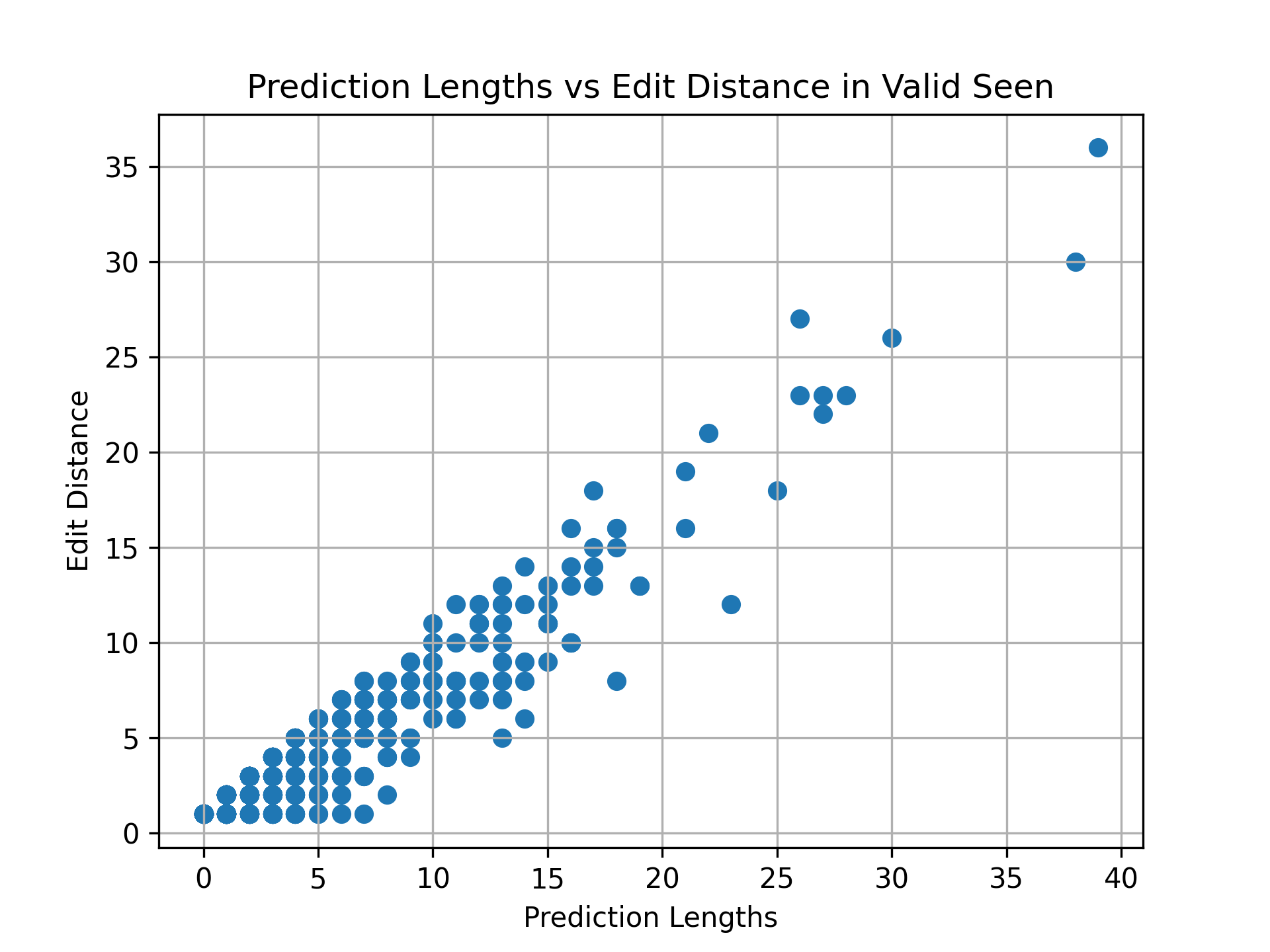} \\
        \includegraphics[scale=0.5]{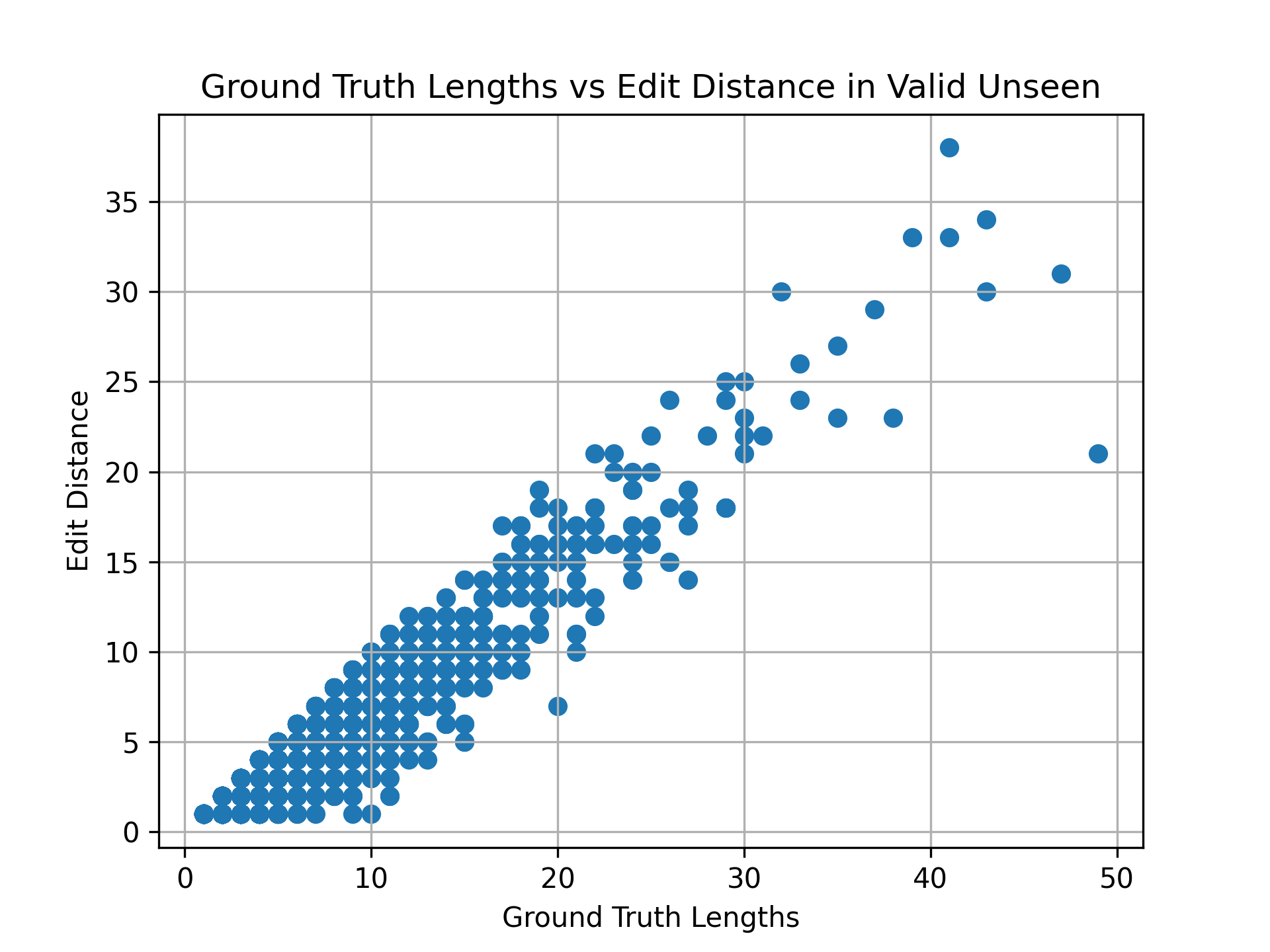} & \includegraphics[scale=0.5]{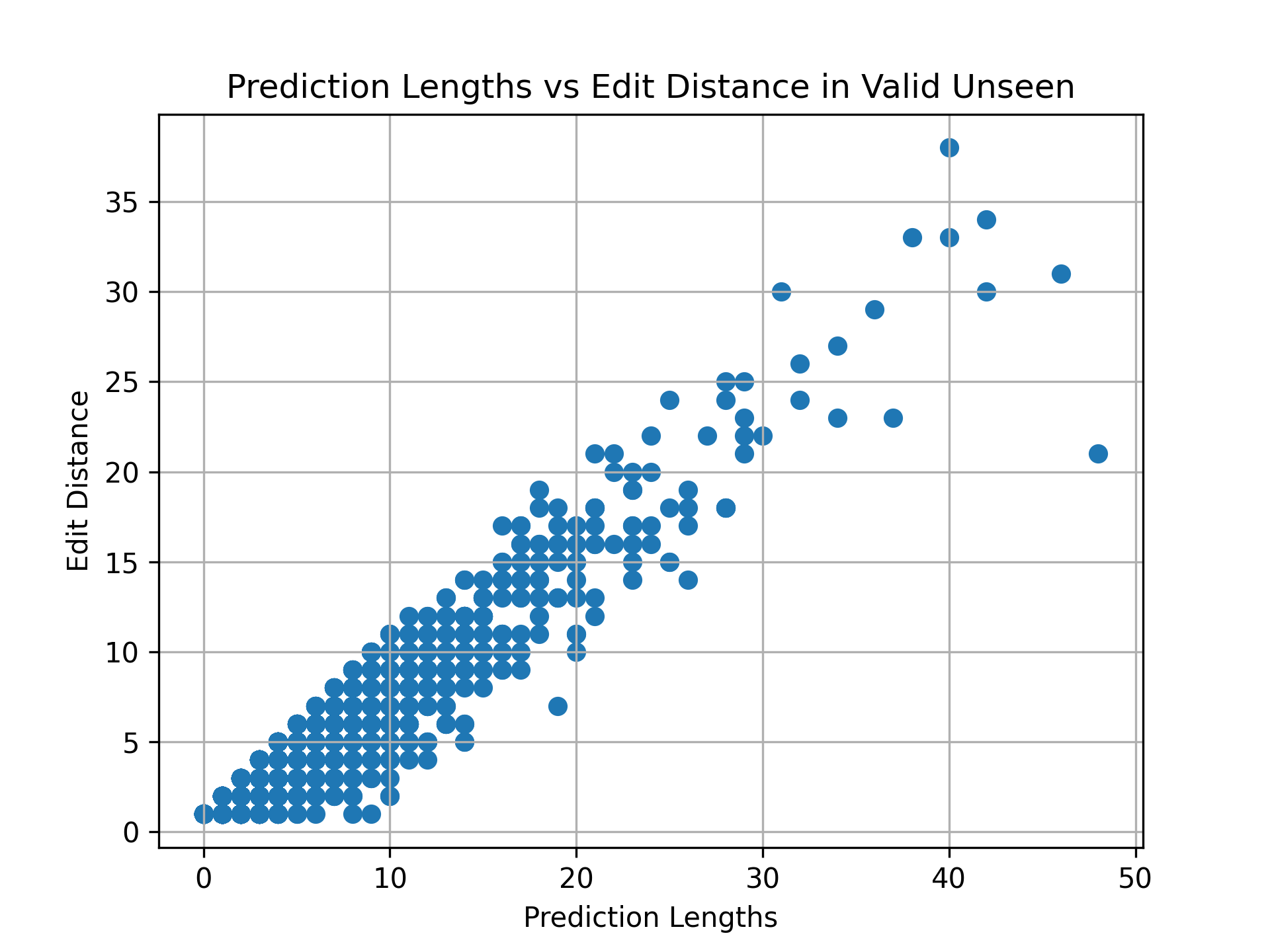}
    \end{tabular}

    \caption{These plots show the distribution of edit distances and the ground truth and prediction lengths. As the length of the ground truth or predicted plan increases, the possibility of having a larger edit distance increases.}
    \label{fig:norm_edit_dist}
\end{figure*}

One of the metrics we use to evaluate predicted plans is edit distance~\cite{editdistance} to the ground truth plan. For computing edit distance, we treat each \texttt{(action, object)} pair as a token and compute the minimum number of edits - insertions, deletions and swaps - needed to convert the predicted plan to the ground truth plan. 
For example, given a ground truth plan \texttt{(Pickup, Mug), (Place, CoffeeMachine), (ToggleOn, CoffeeMachine)}, the following are the edit distances for a few sample predicted plans:
\begin{itemize}
    \item Predicted plan: \texttt{(Pickup, CounterTop), (Place, CoffeeMachine), (ToggleOn, CoffeeMachine)} - Edit distance is 1 since the first step needs a swap of the object (we do not use assign partial credit for predicting the right action).
    \item Predicted Plan: \texttt{(Pickup, Mug), (ToggleOn, CoffeeMachine)} - Edit distance is 1 since the step \texttt{(Place, CoffeeMachine)} needs to be inserted.
    \item Predicted Plan: \texttt{(Pickup, Mug), (Place, CoffeeMachine), (ToggleOn, CoffeeMachine), (ToggleOff, CoffeeMachine)} - Edit distance is 1 since the final strep \texttt{(ToggleOff, CoffeeMachine)} needs to be deleted.
\end{itemize}
We can also see that depending on how task success is defined, the edit distance of a plan may not correlate well with task success. For example, suppose the task of making coffee only involves checking if the mug eventually has coffee, while all the above three plans have an edit distance of 1, the third plan alone has a task success of 1 but the other two will not succeed as both those plans fail to relocate the mug to the coffee machine, in which case it will not get filled with coffee. 

We can see from Figure~\ref{fig:norm_edit_dist} that our distribution of ground truth is skewed towards shorter lengths, and they correlate with the edit distance linearly. This means that edit distance may not be comparable or insightful directly between different models if the output is of different length. We observe such difference between models in Table~\ref{tab:all_metrics_div_val_unseen}, where the baseline has a low edit distance compared to other ET models. In order to make them comparable without relying on the length of the ground truth, we normalize them according to their prediction lengths. This results in comparable results that are on the same scale, and shows the performance in terms of a corrected edit distance.

Prediction Length normalization shows that the edit distance in all methods are very similar to each other. Most are different from the ground truth length. 

\section{Analyzing success rate across tasks}

\begin{table*}[!ht]
\resizebox{\textwidth}{!}{%
    \centering
    \begin{tabular}{L{2cm}ccccccccccccc} 
\toprule
\multirow{2}{*}{Task}& \multicolumn{2}{c}{Baseline}& \multicolumn{2}{c}{E.T.}& \multicolumn{2}{c}{E.T. Hier}& \multicolumn{2}{c}{E.T. Mask}& \multicolumn{2}{c}{Oracle}& \multicolumn{2}{c}{CorefOracle}\\
 & Dir. & Asst. & Dir. & Asst. & Dir. & Asst. & Dir. & Asst. & Dir. & Asst. & Dir. & Asst.\\
\midrule
Coffee & \phantom{0}3.45 & \phantom{0}3.45 & 22.41 & 25.86 & 29.31 & 25.86 & 31.03 & 27.59 & 82.76 & 82.76 & 91.38 & 89.66\\
Water Plant & \phantom{0}0.00 & \phantom{0}0.00 & 53.33 & 76.67 & 56.67 & 63.33 & 70.00 & 73.33 & 83.33 & 86.67 & 93.33 & 93.33\\
Plate Of Toast & 11.11 & 11.11 & \phantom{0}6.67 & 10.00 & 10.00 & 15.56 & \phantom{0}6.67 & 13.33 & 50.00 & 64.44 & 75.56 & 81.11\\
Clean All X & \phantom{0}3.23 & \phantom{0}3.23 & 22.58 & 20.97 & 24.19 & 20.97 & 29.03 & 24.19 & 69.35 & 75.81 & 72.58 & 77.42\\
Put All X On Y & 15.56 & 17.78 & 11.11 & 20.00 & 17.78 & 15.56 & 15.56 & 24.44 & 48.89 & 55.56 & 66.67 & 84.44\\
Put All X In One Y & 16.67 & 18.75 & 18.75 & 22.92 & 14.58 & 18.75 & 20.83 & 25.00 & 62.50 & 70.83 & 77.08 & 79.17\\
N Slices Of X In Y & \phantom{0}7.95 & 11.36 & 15.91 & 22.73 & 10.23 & 15.91 & 14.77 & 17.05 & 60.23 & 65.91 & 75.00 & 78.41\\
N Cooked Slices Of X In Y & 10.29 & 11.76 & 14.71 & 13.24 & 16.18 & 14.71 & 16.18 & 14.71 & 50.74 & 51.47 & 67.65 & 63.24\\
Boil X & 12.50 & 15.00 & 20.00 & 20.00 & 15.00 & 12.50 & 15.00 & 22.50 & 40.00 & 40.00 & 57.50 & 57.50\\
Salad & \phantom{0}5.44 & \phantom{0}8.16 & 12.24 & 14.97 & 13.61 & 14.29 & 12.93 & 12.93 & 44.22 & 52.38 & 63.95 & 64.63\\
Sandwich & \phantom{0}9.29 & 10.71 & 10.71 & 10.71 & \phantom{0}7.86 & 13.57 & 11.43 & 15.00 & 51.43 & 62.14 & 71.43 & 80.00\\
Breakfast & \phantom{0}2.58 & \phantom{0}3.09 & 15.46 & 20.10 & 17.53 & 20.10 & 18.56 & 20.10 & 57.22 & 62.89 & 65.98 & 73.20\\
\bottomrule
    \end{tabular}%
    }
    \caption{Breakdown of success rate by task for split \texttt{divided\_val\_unseen}. Some model names have been shortened due to space limitations. (Dir: Direct, Asst: Assisted)}
    \label{tab:task_level_success}
\end{table*}

We include a breakdown of the success rate of the different models paired with the different execution methods at a task level in table \ref{tab:task_level_success}. By examining oracle performance across tasks, we can confirm that plan execution is more likely to fail for tasks involving more or difficult placement steps. Boiling is the task causing the most oracle failures as it necessitates placing either a bowl in a microwave or a pot on the stove, both of which are challenging in AI2-THOR due to physics constraints. Other tasks resulting in a higher rate of oracle failures are hierarchical tasks such as \texttt{Salad}, \texttt{Sandwich} and \texttt{Breakfast}. 

For the E.T. models, we see consistent increases in task level success rate for most tasks from E.T. to E.T. Hierarchical and E.T. to E.T. + Mask but the trend between E.T. Hierarchical and E.T. + Mask is less consistent. The baseline outperforms vanilla E.T. on some tasks but almost never outperforms E.T. + Hierarchical or E.T. + Mask. We also notice many contrasts between tasks that the baseline is stronger on and tasks that the E.T. models are stronger on. For example in the \texttt{Water Plant} task, the baseline is never successful but the E.T. models perform much better on this than on other tasks. 

We can also see that the benefits of execution assistance, both for models and for the oracles are for tasks involving multiple placement steps, despite the fact that placement is also the object interaction that has the highest failure rate even after assistance.

\section{Qualitative analysis}

\begin{figure*}[!h]
    \centering
    \includegraphics[width=\textwidth]{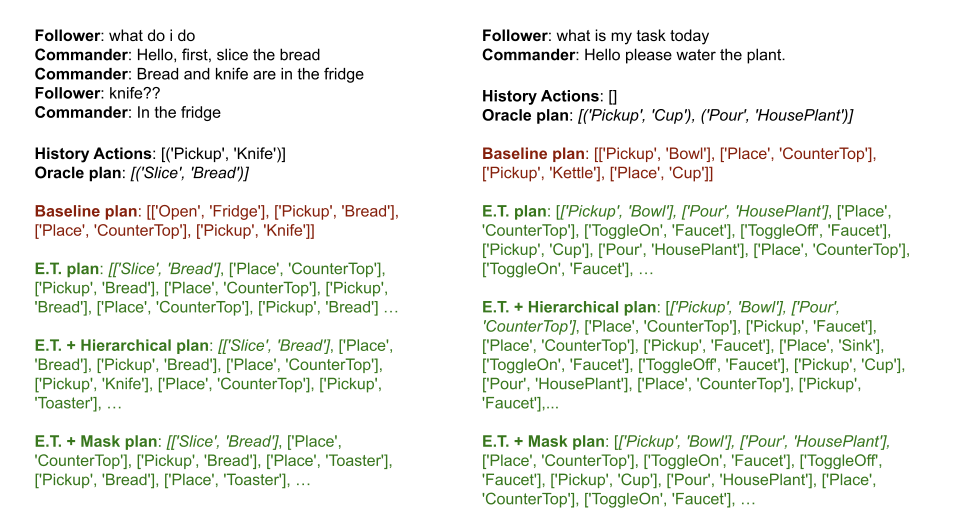}
    \caption{Qualitative examples where the \et\ models outperform the \bart\ baseline. The parts of the predicted plans that complete the necessary state changes are italicized. Multimodal context helps the model better identify how much of the task is already complete, and visual cues appear to help break down tasks requiring additional reasoning. However, the \et\ models also fail to stop after performing the necessary actions.}
    \label{fig:qual_egs_success}
\end{figure*}

\begin{figure*}[!h]
    \centering
    \includegraphics[width=\textwidth]{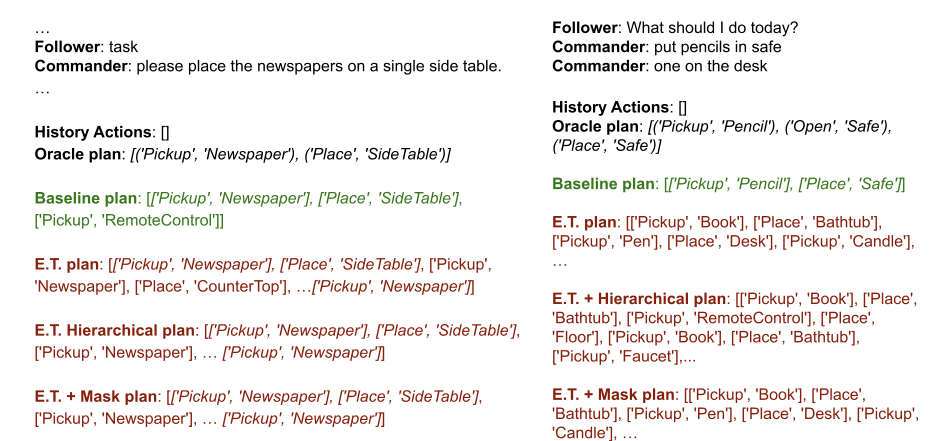}
    \caption{Qualitative examples where the \bart\ baseline outperforms the \et\ models. The parts of the predicted plans that complete the necessary state changes are italicized. These examples demonstrate cases where failing to stop can cause the environment to be in an incorrect state at the end of the episode, and visual cues can be distracting rather than helpful.}
    \label{fig:qual_egs_failure}
\end{figure*}

We include some qualitative examples in figures \ref{fig:qual_egs_success} and \ref{fig:qual_egs_failure}. We can see from the first example in figure \ref{fig:qual_egs_success} that the \et\ models are better able to identify how much of the task discussed in the dialog is already complete by making use of the visual context. The \bart\ plan ends too early before the bread is sliced. In contrast, the \et\ plans start with the right action of slicing bread, but then as they do not learn when to stop, they start performing arbitrary actions on surrounding objects. This issue can lead to failure, as is seen in the first example in figure \ref{fig:qual_egs_failure} where the model reverses the correct actions it initially performed. 

Visual cues also appear to help the \et\ models break down tasks that require more reasoning such as using a water in other utensils to water the plant, as is seen in the second example in figure \ref{fig:qual_egs_success}. However, in some cases they can be distracting, as in the second example in figure \ref{fig:qual_egs_failure} where the model is distracted by larger objects and fails to focus on pencils mentioned in the dialog, which the \bart\ baseline correctly attends to.  
\section{Complete Experimental results}
The complete results with all metrics evaluated for all model and oracle conditions are included in tables \ref{tab:all_metrics_div_val_seen}, \ref{tab:all_metrics_div_val_unseen}, \ref{tab:all_metrics_div_test_seen}, \ref{tab:all_metrics_div_test_unseen} for splits \texttt{divided\_val\_seen}, \texttt{divided\_val\_unseen}, \texttt{divided\_test\_seen} and \texttt{divided\_test\_unseen} respectively. When an oracle condition has a non-zero edit distance, this indicates that one or more oracle trajectories reached the execution failure limit, that is, the trajectory contained 30 plan steps which failed to get executed, at which point the rest of the trajectory is ignored. 

\begin{table*}[!h]
    \centering
    \tabcolsep 2pt
    \begin{tabular}{L{2cm}L{2cm} | C{1cm}C{1cm}C{1.5cm}C{2cm}C{1.5cm}C{1.5cm}}
        \toprule
        Model & Execution & SR & GC & Edit Distance & Frac Valid Plan Steps & GT-Norm ED & Pred-Norm ED \\
        \midrule
\multirow{2}{2cm}{Baseline} & Direct & 11.26 & 13.67 & 4.87 & 94.61 & 1.39 & 1.22\\
 & Assisted & 11.92 & 17.27 & 4.87 & 94.61 & 1.39 & 1.22\\
\multirow{2}{2cm}{E.T.} & Direct & 12.91 & 16.32 & 45.78 & 90.33 & 21.72 & 0.96\\
 & Assisted & 15.89 & 20.57 & 54.80 & 89.08 & 25.48 & 0.96\\
\multirow{2}{2cm}{E.T. Hierarchical} & Direct & 14.24 & 15.67 & 45.08 & 88.25 & 21.49 & 0.96\\
 & Assisted & 18.21 & 20.45 & 51.52 & 86.84 & 23.75 & 0.95\\
\multirow{2}{2cm}{E.T. + Mask} & Direct & 15.23 & 22.51 & 49.43 & 98.51 & 22.82 & 0.96\\
 & Assisted & 18.87 & 28.99 & 60.00 & 98.69 & 27.51 & 0.96\\
\multirow{2}{2cm}{Oracle} & Direct & 61.92 & 63.64 & 0.00 & 85.96 & 0.00 & 0.00\\
 & Assisted & 68.87 & 72.13 & 0.01 & 85.93 & 0.01 & 0.00\\
\multirow{2}{2cm}{CorefOracle} & Direct & 77.81 & 83.03 & 0.01 & 86.13 & 0.00 & 0.00\\
 & Assisted & 80.13 & 84.50 & 0.03 & 86.08 & 0.01 & 0.02\\
        \bottomrule
    \end{tabular}
    \caption{Complete results of different models combined with different execution methods on the \teach\ EDH \texttt{divided\_val\_seen} split.}
    \label{tab:all_metrics_div_val_seen}
\end{table*}

\begin{table*}[!h]
    \centering
    \tabcolsep 2pt
    \begin{tabular}{L{2cm}L{2cm} | C{1cm}C{1cm}C{1.5cm}C{2cm}C{1.5cm}C{1.5cm}}
        \toprule
        Model & Execution & SR & GC & Edit Distance & Frac Valid Plan Steps & GT-Norm ED & Pred-Norm ED \\
        \midrule
\multirow{2}{2cm}{Baseline} & Direct & 7.51 & 11.03 & 5.78 & 95.76 & 1.56 & 1.42\\
 & Assisted & 8.91 & 12.19 & 5.78 & 95.80 & 1.56 & 1.43\\
\multirow{2}{2cm}{E.T.} & Direct & 15.58 & 16.20 & 39.89 & 89.48 & 18.20 & 0.95\\
 & Assisted & 18.74 & 22.36 & 47.21 & 88.78 & 20.83 & 0.95\\
\multirow{2}{2cm}{E.T. Hierarchical} & Direct & 16.23 & 17.27 & 41.91 & 88.93 & 19.24 & 0.94\\
 & Assisted & 18.09 & 24.53 & 49.37 & 88.78 & 22.26 & 0.95\\
\multirow{2}{2cm}{E.T. + Mask} & Direct & 17.81 & 18.29 & 43.48 & 98.79 & 19.68 & 0.95\\
 & Assisted & 19.57 & 27.64 & 52.26 & 98.80 & 22.95 & 0.95\\
\multirow{2}{2cm}{Oracle} & Direct & 55.57 & 58.48 & 0.00 & 85.21 & 0.00 & 0.00\\
 & Assisted & 61.97 & 63.07 & 0.01 & 85.18 & 0.00 & 0.01\\
\multirow{2}{2cm}{CorefOracle} & Direct & 70.87 & 71.87 & 0.13 & 85.72 & 0.01 & 0.02\\
 & Assisted & 74.58 & 77.31 & 0.11 & 85.58 & 0.01 & 0.02\\
        \bottomrule
    \end{tabular}
    \caption{Complete results of different models combined with different execution methods on the \teach\ EDH \texttt{divided\_val\_unseen} split.}
    \label{tab:all_metrics_div_val_unseen}
\end{table*}

\begin{table*}[!h]
    \centering
    \tabcolsep 2pt
    \begin{tabular}{L{2cm}L{2cm} | C{1cm}C{1cm}C{1.5cm}C{2cm}C{1.5cm}C{1.5cm}}
        \toprule
        Model & Execution & SR & GC & Edit Distance & Frac Valid Plan Steps & GT-Norm ED & Pred-Norm ED \\
        \midrule
\multirow{2}{2cm}{Baseline} & Direct & 7.19 & 9.62 & 5.75 & 95.55 & 1.58 & 1.47\\
 & Assisted & 9.80 & 12.30 & 5.75 & 95.55 & 1.58 & 1.47\\
\multirow{2}{2cm}{E.T.} & Direct & 15.03 & 19.52 & 44.10 & 88.79 & 21.18 & 0.95\\
 & Assisted & 16.67 & 19.96 & 52.16 & 88.43 & 24.00 & 0.95\\
\multirow{2}{2cm}{E.T. Hierarchical} & Direct & 14.71 & 17.97 & 43.39 & 89.82 & 19.68 & 0.94\\
 & Assisted & 17.97 & 23.67 & 55.06 & 88.18 & 24.97 & 0.96\\
\multirow{2}{2cm}{E.T. + Mask} & Direct & 16.34 & 23.84 & 49.01 & 99.03 & 22.89 & 0.95\\
 & Assisted & 18.95 & 26.35 & 62.47 & 98.97 & 28.85 & 0.95\\
\multirow{2}{2cm}{Oracle} & Direct & 54.58 & 53.58 & 0.01 & 87.54 & 0.00 & 0.00\\
 & Assisted & 61.44 & 62.49 & 0.00 & 87.55 & 0.00 & 0.00\\
\multirow{2}{2cm}{CorefOracle} & Direct & 75.82 & 79.50 & 0.13 & 87.97 & 0.01 & 0.04\\
 & Assisted & 78.43 & 80.92 & 0.11 & 87.84 & 0.01 & 0.04\\
        \bottomrule
    \end{tabular}
    \caption{Complete results of different models combined with different execution methods on the \teach\ EDH \texttt{divided\_test\_seen} split.}
    \label{tab:all_metrics_div_test_seen}
\end{table*}

\begin{table*}[!h]
    \centering
    \tabcolsep 2pt
    \begin{tabular}{L{2cm}L{2cm} | C{1cm}C{1cm}C{1.5cm}C{2cm}C{1.5cm}C{1.5cm}}
        \toprule
        Model & Execution & SR & GC & Edit Distance & Frac Valid Plan Steps & GT-Norm ED & Pred-Norm ED \\
        \midrule
\multirow{2}{2cm}{Baseline} & Direct & 8.87 & 9.54 & 5.46 & 94.99 & 1.45 & 1.34\\
 & Assisted & 10.27 & 12.31 & 5.45 & 95.03 & 1.45 & 1.35\\
\multirow{2}{2cm}{E.T.} & Direct & 16.62 & 15.61 & 41.37 & 89.88 & 18.74 & 0.95\\
 & Assisted & 19.98 & 27.13 & 47.36 & 89.30 & 20.71 & 0.95\\
\multirow{2}{2cm}{E.T. Hierarchical} & Direct & 17.27 & 20.30 & 41.67 & 88.69 & 18.90 & 0.95\\
 & Assisted & 19.70 & 25.82 & 48.44 & 88.18 & 20.93 & 0.95\\
\multirow{2}{2cm}{E.T. + Mask} & Direct & 17.46 & 18.96 & 44.80 & 99.02 & 20.27 & 0.95\\
 & Assisted & 20.07 & 28.33 & 53.80 & 98.98 & 23.29 & 0.96\\
\multirow{2}{2cm}{Oracle} & Direct & 56.77 & 58.01 & 0.00 & 85.77 & 0.00 & 0.00\\
 & Assisted & 63.21 & 64.87 & 0.02 & 85.71 & 0.01 & 0.01\\
\multirow{2}{2cm}{CorefOracle} & Direct & 71.90 & 74.34 & 0.14 & 86.17 & 0.01 & 0.02\\
 & Assisted & 76.94 & 78.30 & 0.13 & 86.10 & 0.01 & 0.02\\
        \bottomrule
    \end{tabular}
    \caption{Complete results of different models combined with different execution methods on the \teach\ EDH \texttt{divided\_test\_unseen} split.}
    \label{tab:all_metrics_div_test_unseen}
\end{table*}

\section{FILM-like model for TEACh EDH task}

Consider the following sample gameplay session from the TEACh dataset. In the video, the top-left panel is the first person view that the Follower (Robot) would see. Below are just the utterances interspersed with compressed action sequences where every sequence of navigation actions is just replaced by a generic “navigate” action.

\begin{lstlisting}
Game file: c7d8b93c8a4486f3_8e32.game.json 
Task: N Cooked Slices Of X In Y 
Params: ['2', 'Potato', 'on', 'Plate']
Resolved task description: Serve 2 cooked slices of potato on a plate

Driver : what shall we do today?
Commander : make 2 sliceses of potato

Driver actions: Navigate Potato|-02.99|+00.75|+02.39 -- Pickup Potato|-02.99|+00.75|+02.39

Commander : potatao is inside the sink

Driver : Navigate DiningTable|-00.49|00.00|+03.18 -- Place Potato on DiningTable|-00.49|00.00|+03.18 -- 
Pickup ButterKnife|-00.70|+00.87|+03.43 -- Navigate Potato|-02.99|+00.75|+02.39 -- 
Slice Potato|-02.99|+00.75|+02.39

Driver : i cut potato

Driver : Place Knife on DiningTable|-00.49|00.00|+03.18

Driver : what shall i do next
Commander : cook 2 slices potato in micro wave

Driver : Navigate Potato|-02.99|+00.75|+02.39|PotatoSliced_0 -- 
Pickup Potato|-02.99|+00.75|+02.39|PotatoSliced_0 -- Navigate CounterTop|-01.37|+00.95|+00.35 -- 
Place potato slice on CounterTop|-01.37|+00.95|+00.35 -- Open Microwave|-00.36|+00.92|+00.22 -- 
Pickup Apple|-00.30|+01.07|+00.14 -- Place apple in Sink|-01.11|+00.93|+00.34|SinkBasin -- 
Navigate Potato|-02.99|+00.75|+02.39|PotatoSliced_5 -- 
Pickup Potato|-02.99|+00.75|+02.39|PotatoSliced_5 -- Navigate Microwave|-00.36|+00.92|+00.22 -- 
Place Microwave|-00.36|+00.92|+00.22 -- Pickup Fork|-00.24|+01.02|+00.27 -- 
Place fork in Sink|-01.11|+00.93|+00.34|SinkBasin -- Pickup 
Potato|-02.99|+00.75|+02.39|PotatoSliced_0 -- Place potato slice in Microwave|-00.36|+00.92|+00.22 -- 
Close Microwave|-00.36|+00.92|+00.22 -- ToggleOn Microwave|-00.36|+00.92|+00.22 -- 
Navigate unk_obj-- Open unk_obj -- Open unk_obj -- Navigate unk_obj -- Open unk_obj -- 
Navigate Fridge|-02.80|+00.00|+04.43 -- Open Fridge|-02.80|+00.00|+04.43

Commander : plate is inside the fridge

Driver : Navigate Cup|-02.81|+01.19|+04.28 -- Pickup Cup|-02.81|+01.19|+04.28 -- 
Place cup in Fridge|-02.80|+00.00|+04.43 -- Pickup Plate|-02.81|+01.18|+04.28 -- 
Place plate on CounterTop|-01.37|+00.95|+00.35 -- ToggleOff Microwave|-00.36|+00.92|+00.22 -- 
Open Microwave|-00.36|+00.92|+00.22 -- 
Pickup Potato|-02.99|+00.75|+02.39|PotatoSliced_0 -- Navigate Plate|-02.81|+01.18|+04.28 -- 
Place potato slice on Plate|-02.81|+01.18|+04.28 -- 
Navigate Potato|-02.99|+00.75|+02.39|PotatoSliced_5 -- 
Pickup Potato|-02.99|+00.75|+02.39|PotatoSliced_5 -- 
Navigate Plate|-02.81|+01.18|+04.28 -- Place Plate|-02.81|+01.18|+04.2
\end{lstlisting}

In the EDH task we would get the dialog an action history until some break in the dialog and have to predict actions until the next dialog utterance occurs. \\ 

Sample EDH instance 1:

\begin{lstlisting}
Dialog History:
Driver : what shall we do today?
Commander : make 2 sliceses of potato
Commander : potatao is inside the sink
Driver : i cut potato
Driver : what shall i do next
Commander : cook 2 slices potato in micro wave

Driver action history: 
Navigate Potato|-02.99|+00.75|+02.39 -- Pickup Potato|-02.99|+00.75|+02.39
Navigate DiningTable|-00.49|00.00|+03.18 -- Place Potato on DiningTable|-00.49|00.00|+03.18 -- 
Pickup ButterKnife|-00.70|+00.87|+03.43 -- Navigate Potato|-02.99|+00.75|+02.39 -- 
Slice Potato|-02.99|+00.75|+02.39
Place Knife on DiningTable|-00.49|00.00|+03.18

Actions to be predicted: 
Navigate Potato|-02.99|+00.75|+02.39|PotatoSliced_0 -- 
Pickup Potato|-02.99|+00.75|+02.39|PotatoSliced_0 -- Navigate CounterTop|-01.37|+00.95|+00.35 -- 
Place potato slice on CounterTop|-01.37|+00.95|+00.35 -- Open Microwave|-00.36|+00.92|+00.22 -- 
Pickup Apple|-00.30|+01.07|+00.14 -- Place apple in Sink|-01.11|+00.93|+00.34|SinkBasin -- 
Navigate Potato|-02.99|+00.75|+02.39|PotatoSliced_5 -- 
Pickup Potato|-02.99|+00.75|+02.39|PotatoSliced_5 -- Navigate Microwave|-00.36|+00.92|+00.22 -- 
Place Microwave|-00.36|+00.92|+00.22 -- Pickup Fork|-00.24|+01.02|+00.27 -- 
Place fork in Sink|-01.11|+00.93|+00.34|SinkBasin -- Pickup 
Potato|-02.99|+00.75|+02.39|PotatoSliced_0 -- Place potato slice in Microwave|-00.36|+00.92|+00.22 -- 
Close Microwave|-00.36|+00.92|+00.22 -- ToggleOn Microwave|-00.36|+00.92|+00.22 -- 
Navigate unk_obj-- Open unk_obj -- Open unk_obj -- Navigate unk_obj -- Open unk_obj -- 
Navigate Fridge|-02.80|+00.00|+04.43 -- Open Fridge|-02.80|+00.00|+04.43

Success criteria: 2 potato slices get cooked
\end{lstlisting}

Sample EDH instance 2:

\begin{lstlisting}
Dialog History:
Driver : what shall we do today?
Commander : make 2 sliceses of potato
Commander : potatao is inside the sink
Driver : i cut potato
Driver : what shall i do next
Commander : cook 2 slices potato in micro wave
Commander : plate is inside the fridge

Driver action history: 
Navigate Potato|-02.99|+00.75|+02.39 -- Pickup Potato|-02.99|+00.75|+02.39
Navigate DiningTable|-00.49|00.00|+03.18 -- Place Potato on DiningTable|-00.49|00.00|+03.18 -- 
Pickup ButterKnife|-00.70|+00.87|+03.43 -- Navigate Potato|-02.99|+00.75|+02.39 -- 
Slice Potato|-02.99|+00.75|+02.39
Place Knife on DiningTable|-00.49|00.00|+03.18
Navigate Potato|-02.99|+00.75|+02.39|PotatoSliced_0 -- 
Pickup Potato|-02.99|+00.75|+02.39|PotatoSliced_0 -- Navigate CounterTop|-01.37|+00.95|+00.35 -- 
Place potato slice on CounterTop|-01.37|+00.95|+00.35 -- Open Microwave|-00.36|+00.92|+00.22 -- 
Pickup Apple|-00.30|+01.07|+00.14 -- Place apple in Sink|-01.11|+00.93|+00.34|SinkBasin -- 
Navigate Potato|-02.99|+00.75|+02.39|PotatoSliced_5 -- 
Pickup Potato|-02.99|+00.75|+02.39|PotatoSliced_5 -- Navigate Microwave|-00.36|+00.92|+00.22 -- 
Place Microwave|-00.36|+00.92|+00.22 -- Pickup Fork|-00.24|+01.02|+00.27 -- 
Place fork in Sink|-01.11|+00.93|+00.34|SinkBasin -- Pickup 
Potato|-02.99|+00.75|+02.39|PotatoSliced_0 -- Place potato slice in Microwave|-00.36|+00.92|+00.22 -- 
Close Microwave|-00.36|+00.92|+00.22 -- ToggleOn Microwave|-00.36|+00.92|+00.22 -- 
Navigate unk_obj-- Open unk_obj -- Open unk_obj -- Navigate unk_obj -- Open unk_obj -- 
Navigate Fridge|-02.80|+00.00|+04.43 -- Open Fridge|-02.80|+00.00|+04.43

Actions to be predicted: 
Navigate Cup|-02.81|+01.19|+04.28 -- Pickup Cup|-02.81|+01.19|+04.28 -- 
Place cup in Fridge|-02.80|+00.00|+04.43 -- Pickup Plate|-02.81|+01.18|+04.28 -- 
Place plate on CounterTop|-01.37|+00.95|+00.35 -- ToggleOff Microwave|-00.36|+00.92|+00.22 -- 
Open Microwave|-00.36|+00.92|+00.22 -- 
Pickup Potato|-02.99|+00.75|+02.39|PotatoSliced_0 -- Navigate Plate|-02.81|+01.18|+04.28 -- 
Place potato slice on Plate|-02.81|+01.18|+04.28 -- 
Navigate Potato|-02.99|+00.75|+02.39|PotatoSliced_5 -- 
Pickup Potato|-02.99|+00.75|+02.39|PotatoSliced_5 -- 
Navigate Plate|-02.81|+01.18|+04.28 -- Place Plate|-02.81|+01.18|+04.2

Success Criteria: Environment contains a plate that contains 2 cooked slices of potato
\end{lstlisting}

I will now walk through what it would take to build a FILM (https://arxiv.org/pdf/2110.07342.pdf)-like model for this task and what are the new challenges for TEACh: \\

\textbf{Step 1}: Predict the task and task params from provided dialog history
Challenges:
\begin{enumerate}
    \item TEACh tasks are hierarchical. For example we have both a Make Coffee task and a Prepare Breakfast task that has Make Coffee as a subtask. Sometimes the dialog history will only mention the subtask and not the full task. Update: Some preliminary experiments at Amazon suggest that task prediction is pretty simple.
    \item The early part of the dialog may not reference all task params. In the above example, the plate is only mentioned after the potato slices are cooked. If the EDH instance we have to predict requires predicting some of the earlier actions, our method needs to be robust to the fact that the plate is currently unknown.
\end{enumerate}

\textbf{Step 2}: Retrieve a hand-crafted plan template for the task and fill it in with params
Challenges:
\begin{enumerate}
    \item We can create task templates at a task level but we may prefer to do it at the level of desired object state changes. 
    \item For the EDH task we only have to predict some specific subsequence of actions. Hence we only need part of the hand-crafted plan. So we need to detect what part of the plan has already been completed from the action history.
    \item Dialog history can mention actions to be performed at different levels of abstraction 
    \item In ALFRED, all necessary objects are somewhere visible and so it is possible to walk around the room and find all of them. In TEACh we likely need to place more emphasis on understanding any information the dialog provides about locations of objects
    \item Plans will involve looping (eg: for tasks involving "all" like place all forks in the sink) and if-then (eg: turn off the tap if it is on before placing something in the sink) 
\end{enumerate}

\textbf{Step 3}: Build a semantic map of the environment (should be same as ALFRED)

\textbf{Step 4}: Implement a navigation / object search policy
Challenges:
\begin{enumerate}
    \item If object location information is not included in plan, object search policy will need to use relevant utterances in dialog
    \item The ground truth action sequence may use another instance of an object type placed at a different location from the one suggested in the dialog
\end{enumerate}

\section{Tentative Weekly Plan}
\begin{itemize}
    \item Weeks 1-2:  Onboarding; Set up internal TEACh codebase; Get familiar with baseline models for EDH, and classification of tasks, task parameters and plans from dialog
    \item Weeks 3-5: Work on improving prediction of desired object manipulations. Possible strategies
    \begin{itemize}
        \item Design a classification based model for predicting task params from dialog + Implement programmatic method to convert task definitions into plans + Identifying completed steps from past actions
        \item Explore alternative problem formulations for directly predicting object manipulations from dialog 
    \end{itemize}
    \item Weeks 6-8: Incorporate object locations into plans
    \item Weeks 9-10: Implement basic execution of plans using assistive ground truth navigation hook
    \item Week 11: 
    \begin{itemize}
        \item Evaluate end-to-end EDH performance with ground truth navigation
        \item Identify common causes of failure
    \end{itemize}
    \item Week 12: Code cleanup; knowledge transfer; final presentation and document / paper writing 
\end{itemize}

\end{document}